# Multi-Stage Evolutionary Model Merging with Meta Data Driven Curriculum Learning for Sentiment-Specialized Large Language Modeling

Keito Inoshita
Faculty of Data Science
Shiga University, Hikone, Japan
Email: inosita.2865@gmail.com

Xiaokang Zhou
Faculty of Business Data Science
Kansai University, Osaka, Japan
RIKEN Center for AIP
Tokyo, Japan
Email: zhou@kansai-u.ac.jp

Akira Kawai
Faculty of Data Science
Shiga University, Hikone, Japan
Japan Safety Society Research Center
Email: Akira-kawai@biwako.shiga-u.ac.jp

*Abstract*—The emergence of large language models (LLMs) has significantly transformed natural language processing (NLP), enabling more generalized models to perform various tasks with minimal training. However, traditional sentiment analysis methods, which focus on individual tasks such as sentiment classification or aspect-based analysis, are not practical for real-world applications that usually require handling multiple tasks. While offering flexibility, LLMs in sentiment-specific tasks often fall short of the required accuracy. Techniques like fine-tuning and evolutionary model merging help integrate models into a unified framework, which can improve the learning performance while reducing computational costs. The use of task meta-data and curriculum learning to optimize learning processes remains underexplored, while sentiment analysis is a critical task in NLP that requires high accuracy and scalability across multiple subtasks. In this study, we propose a hybrid learning model called Multi-stage Evolutionary Model Merging with Meta data driven Curriculum Learning (MEM-MCL), to enhance the sentiment analysis in large language modeling. In particular, expert models are created through instruction tuning for specific sentiment tasks and then merged using evolutionary algorithms to form a unified model. The merging process is optimized with weak data to enhance performance across tasks. The curriculum learning is incorporated to provide a learning sequence based on task difficulty, improving knowledge extraction from LLMs. Experiment results demonstrate that the proposed MEM-MCL model outperforms conventional LLMs in a majority of sentiment analysis tasks, achieving superior results across various subtasks.

## I. Introduction

The advent of large language models (LLMs) has transformed natural language processing (NLP), replacing many task-specific models with more flexible and generalized systems. LLMs have excelled across tasks with minimal additional training, revolutionizing areas such as machine translation and question answering. Sentiment analysis, as a key NLP task, has benefited from deep learning advancements that enable better detection of subtle emotions in text, improving accuracy and efficiency [1]. However, traditional methods often focus on individual tasks, making them less practical for real-world applications where multiple tasks, such as sentiment classification and aspect-based analysis, must be handled simultaneously. These approaches may lack scalability, limiting their use in broader implementations. LLMs, with their flexibility, provide a promising solution but often fall short in high-accuracy sentiment analysis [2]. Supervised fine-tuning (SFT) and in-context learning (ICL) have been applied to improve LLMs' performance, but results remain suboptimal [3, 4]. Additionally, the large size of LLMs presents operational challenges. One solution is model merging, which integrates multiple models into a single framework, improving performance while maintaining a fixed model size. Evolutionary model merging techniques can achieve effective merging without GPU resources, making them highly efficient [5]. This allows LLMs to perform well across multiple tasks without the inefficiencies of separate models.

Despite these advancements, there are still challenges to address. Current approaches often overlook the potential of task meta-data, which play a crucial role in helping LLMs better understand and differentiate between various tasks. While traditional deep learning methods may not fully leverage this information, task meta-data could significantly enhance LLM performance by providing structured insights that improve task generalization. Additionally, curriculum learning, which presents tasks in increasing order of difficulty, can further optimize learning processes. This structured approach, particularly when combined with ICL, could allow LLMs to gradually acquire knowledge and tackle more complex sentiment analysis tasks more effectively.

In this paper, aims at improving sentiment analysis across a wide range of tasks, we propose a Multi-stage Evolutionary Model merging with Meta data driven Curriculum Learning (MEM-MCL) model for Sentiment-Specialized Large Language Modeling, which integrates task-specific expert models through evolutionary merging, while leveraging task meta data and curriculum learning to enhance the model's knowledge extraction through ICL. In particular, our MEM-MCL is designed to: i) improve accuracy across a wide range of sentiment analysis tasks by utilizing a single unified model, addressing the inefficiencies of maintaining separate models for each task; ii) perform effective merging of expert models tailored through instruction tuning by applying MEM, which progressively merges task-specialized models based on task-specific information, thus enhancing performance while maintaining task specialization; and iii) enhance knowledge extraction by employing MCL, which combines curriculum learning with ICL, using task meta data to progressively learn tasks in increasing order of difficulty, resulting in the improved accuracy and task comprehension. Compared with traditional approaches, the proposed MEM-MCL integrates specialized expert models using MEM, maintaining task-specific performance while reducing the need for multiple models, thus improving scalability and efficiency.

Additionally, MCL enables structured learning, allowing smooth transitions from simple to complex tasks based on task meta data. This combination enhances both the efficiency and accuracy of sentiment analysis tasks.

The main contributions of this study are as follows.
i) An integrated framework is designed to enhance the accuracy of sentiment analysis across a wide range of tasks, which combines the curriculum learning with task-specific expertise, to efficiently handle diverse sentiment analysis tasks while maintaining high performance and scalability in complex environments.
ii) A MEM mechanism is developed to utilize instruction tuning to progressively create expert models specialized for each sentiment analysis task, in which expert models are merged using an evolutionary algorithm based on task-specific information, ensuring performance improvements across multiple tasks while retaining task-specific expertise.
iii) A MCL mechanism is devised to employ task meta data to structure learning based on task difficulty, which integrates curriculum learning with ICL to enhance the model's ability in extracting relevant knowledge, while improving accuracy and enabling the model to handle both simple and complex sentiment analysis tasks more effectively.

The rest of this paper is structured as follows. Section II reviews related works on model merging and curriculum learning. Section III introduces the proposed model and the corresponding mechanisms. Section IV demonstrates experiment evaluation, analysis, and discussions, followed by the conclusion in Section V.

## II. RELATED WORKS

### A. Model Merging Methods for LLMs

The rapid advancement of LLMs in NLP, with models including GPT [6] and Gemini [7], has led to high performance across a broad range of tasks. Among these, Llama [8] standed out as an open-source model, favored for its ease of fine-tuning, making it a popular choice for researchers. As LLMs developed, their use in sentiment analysis is also growing. Bubeck et al. [9] highlighted that sentiment analysis, particularly understanding human emotions, is a key for advancing AI, with LLMs playing a pivotal role. Deng et al. [10] applied zero-shot learning to sentiment analysis, demonstrating LLMs' effectiveness in basic tasks, though they often underperformed in more complex tasks such as ABSA and MAST compared to smaller, specialized models.

To address the diverse needs of sentiment analysis tasks, model merging methods used in other domains show promise. Jin et al. [11] proposed merging the weights of language models from different domains without additional data, creating a single model with strong performance across multiple areas—an approach that could benefit sentiment analysis. Similarly, Li et al. [12] introduced the Branch-Train-Merge method, which allowed for parallel training on different data subsets, yielding an effective multi-domain model. Xiao et al. [13] presented the LM-Cocktail method, which merged fine-tuned LLMs to handle both specific and general tasks. These approaches suggested that merging models from different domains can enhance performance in sentiment analysis. Recently, Akiba et al. introduced a merging method based on evolutionary algorithms that optimizes model weights without needing extra data or large computational resources, further advancing the field.

### B. Curriculum Learning for Sentiment Analysis

Sentiment analysis is a crucial technique for understanding human emotions and responding accordingly, widely studied in the field of NLP. Early sentiment analysis methods, as proposed by Taboada [14], relied on dictionary-based approaches, which used sentiment lexicons to predict emotions in text. However, this approach required extensive manual effort to create dictionaries and could not handle unknown words. With the introduction of deep learning, the accuracy of sentiment analysis improved significantly. Abdullah and Ahmet [15] demonstrated that the shift towards transformer models in sentiment analysis, particularly in comparison to traditional recurrent neural networks (RNNs) and convolutional neural networks (CNNs), has shown superiority. For Aspect-Based Sentiment Analysis (ABSA), machine learning methods proposed by Wang et al. [16] were widely used, but more recently, Mohammadi and Shaverizade [17] have proposed an ensemble approach combining CNN, LSTM, BiLSTM, and GRU models to further improve prediction accuracy. Additionally, He et al. [18] introduced a meta-based self-learning approach for ABSA, improving accuracy with limited labeled data, and balancing learning across tasks.

While these methods focused on specific tasks, they struggled to process multiple tasks with high accuracy simultaneously. To address this issue, Sido and Konopík [19] introduced curriculum learning for text data, reorganizing datasets from simple to complex examples to improve model learning efficiency. Rao et al. [20] proposed a curriculum learning strategy leveraging SentiWordNet, which enhanced the performance of sentiment analysis models. Dahiya et al. [21] applied curriculum learning to code-mixed Hindi and English text, aiming to improve sentiment analysis accuracy in multilingual environments. Previous research largely focused on curriculum learning based on the emotional aspects of the data itself, such as word-level sentiment intensity. However, there is a lack of curriculum methods that focus on the meta-information of the tasks themselves. Curriculum learning that utilizes meta-information from sentiment analysis tasks is crucial for achieving consistent high accuracy across multiple tasks in LLM learning. By designing curricula that consider task difficulty and relationships based on meta-information, models can effectively adapt to various sentiment analysis tasks. This approach is expected to enhance the accuracy of sentiment analysis even in general-purpose models.

## III. MULTI-STAGE EVOLUTIONARY MODEL MERGING WITH META DATA DRIVEN CURRICULUM LEARNING

### A. Framework Overview

The architecture of MEM-MCL is specifically designed to enhance the accuracy of sentiment analysis tasks across a wide range of domains. Fig. 1 illustrates the detailed architecture, which synergistically combines the MEM mechanism for expert model integration and the MCL mechanism for efficient knowledge extraction, which aims to leverage the strengths of both components to improve model performance and adaptability for diverse sentiment analysis tasks. The MEM mechanism is responsible for merging task-specific expert models, which can be fine-tuned through instruction tuning. These models are progressively integrated using an evolutionary algorithm that optimizes the merging process based on task-specific meta data. This approach enhances the generalization capabilities of the model while maintaining task-specific performance. By leveraging weak data—instances where individual expert models struggle—MEM fosters mutual complementarity, allowing the model to address weaknesses and improve overall accuracy. This integration process is particularly useful in optimizing model performance for sentiment analysis tasks that require both flexibility and precision. Building on this, the MCL mechanism is designed to further refine the model's learning process by introducing a curriculum learning approach. MCL structures the learning process based on task difficulty, as defined by the task meta data, which allows the model to sequentially progress from simpler tasks to more complex ones, improving its understanding and ability to generalize across various sentiment analysis tasks. The combination of curriculum learning with ICL ensures that the model gradually acquires knowledge in a structured manner, making it more effective in handling a range of tasks from basic sentiment classification to more intricate aspect-based sentiment analysis.

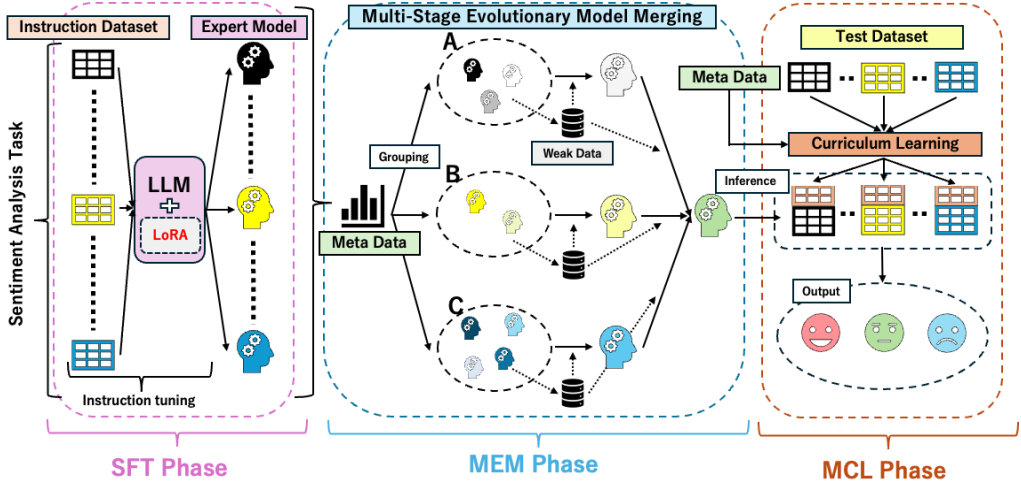

Fig. 1. Framework Overview of MEM-MCL.

In summary, MEM-MCL integrates expert model merging with task-specific curriculum learning, representing a significant advancement in sentiment analysis. By leveraging the strengths of MEM and MCL, the proposed framework ensures that the model can handle a wide array of sentiment analysis tasks with high accuracy, while efficiently extracting and synthesizing knowledge through a multi-staged process.

*B. Sentiment Analysis Tasks and Meta-Information*

Before detailing the proposed method, we explain the sentiment analysis tasks and their associated meta-information. Based on [22], we classify sentiment analysis tasks into three main categories and utilize the associated tasks and meta-information. Fig. 2 presents an overview of these tasks.

The first task is Sentiment Classification (SC), which assigns simple emotion labels, e.g., "Positive" or "Negative" to text, and its difficulty is relatively low. SC tasks include binary classification, three-class classification adding "Neutral," and five-class classification with "Very Positive" and "Very Negative." These tasks play a foundational role in sentiment analysis and are among the most widely studied. Meta-information such as the number of classifications, data diversity, and task simplicity are considered here. Next is Aspect-Based Sentiment Analysis (ABSA), which links emotions to specific aspects or categories within a text, requiring hierarchical thinking. ABSA includes tasks like Aspect Term Sentiment Analysis (ATSA) and Aspect Category Sentiment Analysis (ACSA), as well as Targeted Sentiment Detection (TSD) and Aspect Sentiment Detection (ASD), which extract aspects or categories alongside emotions. More advanced tasks, such as Aspect Sentiment Triplet Extraction (ASTE) and Aspect Sentiment Quad Prediction (ASQP), involve simultaneously extracting aspects, sentiments, and opinions. In these tasks, the complexity of the task, the requirement for hierarchical thinking, and the need for specific pairings or output formats are critical meta-information. Lastly, The Multifaceted Analysis of Subjective Text (MAST) classifies more nuanced emotions or subjectivity, such as "fun," "sad," or "surprise," into 11 or 28 classes, including multi-label classifications. Additionally, it involves analyzing texts containing meanings such as "hate," "irony," or "offensive." These tasks, involving subjectivity, are challenging even for humans, presenting significant difficulties for models as well. It is noted that meta-information related to the subjective nature of the tasks and the complexity of emotions is used in this context.

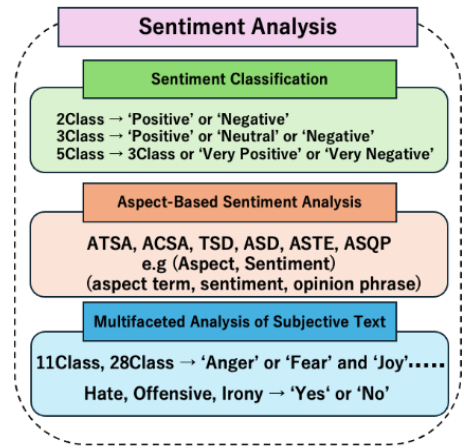

Fig. 2. Grouping of Sentiment Analysis Tasks Based on Meta-Information.

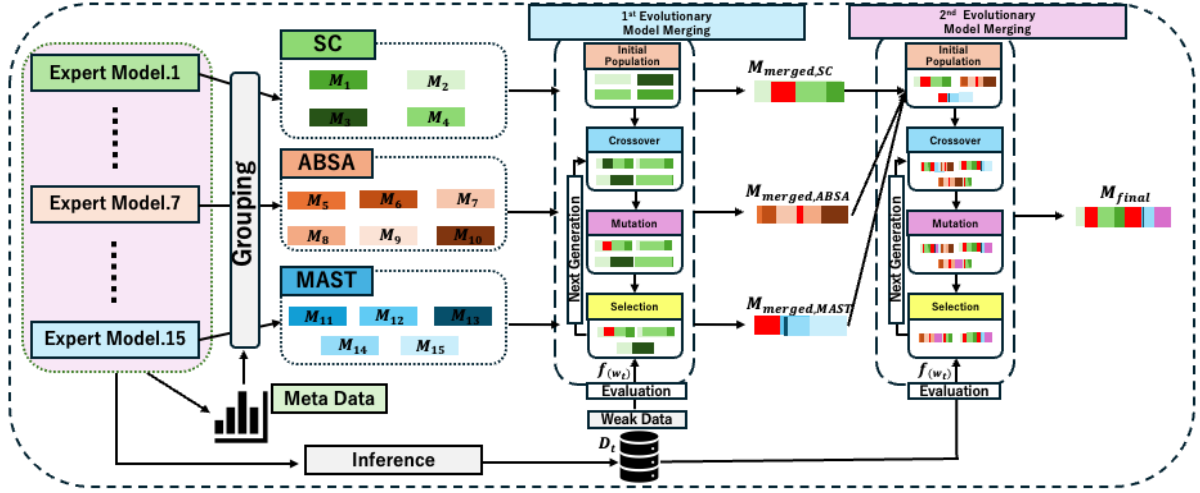

Fig. 4. Mechanism of MEM.

*C. Instruction Tuning for Expert Model*

This section provides a detailed explanation of the SFT phase. In this phase, prompts are created for instruction tuning for each sentiment analysis task, and LoRA [23] is applied to the model to train and build expert models. The process is outlined as follows. First, for each task *t* and data point *i*, the instruction $R_t$, input text $I_{t,i}$, and aspect or category $A_{t,i}$ for ABSA tasks are combined using the Prompt function to generate a prompt $P_{t,i}$. Thus, different prompts are generated for each data point. The prompt creation is expressed as follows.

$$P_{t,i} = Prompt(R_t, I_{t,i}, A_{t,i}) \quad (1)$$

The prompt $P_{t,i}$ is then fed into the LLM with LoRA applied, producing the output $\hat{O}_{t,i}$:

$$\hat{O}_{t,i} = LLM_{RoLA}(P_{t,i}; \theta_t) \quad (2)$$

where $\theta_t$ represents the model parameters, and the goal is to make the predicted output $\hat{O}_{t,i}$ closer to the correct output $O_{t,i}$.

During this training process, the cross-entropy loss function $L_t(\theta_t)$ is used to minimize the error between the predicted output $\hat{O}_{t,i}$ and the ground truth label $O_{t,i}$. The loss function is defined as follows.

$$L_t(\theta_t) = -\sum_i O_{t,i} \log \hat{O}_{t,i} \quad (3)$$

By optimizing this loss function, the model parameters $\theta_t$ are updated to enable the expert model to respond accurately to the specific sentiment analysis task. The algorithm summarizing the model training process is shown in Fig. 3. Based on this training process, expert models specialized for each sentiment analysis task are completed. These expert models will be used in the subsequent merging phase to help construct a generalized model capable of handling multiple tasks.

*D. Evolutionary Algorithm for Multi-Stage Model Merging*

The expert models created during instruction tuning in the SFT phase are merged using the MEM process. The mechanism is illustrated in Fig. 4. First, each expert model is used to infer the data it was trained on, and data points where the model made incorrect predictions, or "weak data" $D_t$, are extracted. This weak data is a critical component of the MEM process.

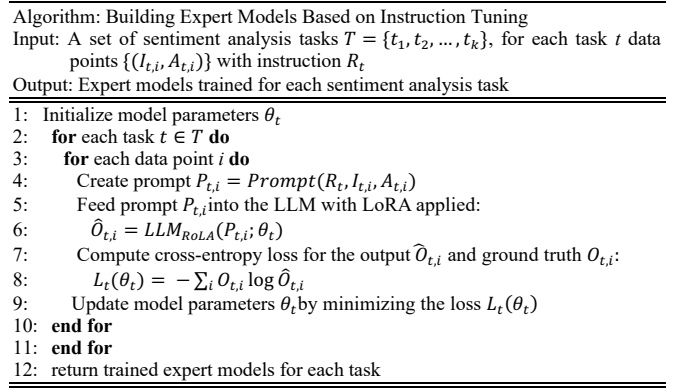

Algorithm: Building Expert Models Based on Instruction Tuning
Input: A set of sentiment analysis tasks $T = \{t_1, t_2, ..., t_k\}$, for each task $t$ data points $\{(I_{t,i}, A_{t,i})\}$ with instruction $R_t$
Output: Expert models trained for each sentiment analysis task

1: Initialize model parameters $\theta_t$
2:   for each task $t \in T$ do
3:     for each data point $i$ do
4:       Create prompt $P_{t,i} = Prompt(R_t, I_{t,i}, A_{t,i})$
5:       Feed prompt $P_{t,i}$ into the LLM with LoRA applied:
6:       $\hat{O}_{t,i} = LLM_{RoLA}(P_{t,i}; \theta_t)$
7:       Compute cross-entropy loss for the output $\hat{O}_{t,i}$ and ground truth $O_{t,i}$:
8:       $L_t(\theta_t) = -\sum_i O_{t,i} \log \hat{O}_{t,i}$
9:       Update model parameters $\theta_t$ by minimizing the loss $L_t(\theta_t)$
10:     end for
11:   end for
12: return trained expert models for each task

Fig. 3. Algorithm for Building Expert Models Based on Instruction Tuning.

This study involves 15 sentiment analysis tasks, which are divided into three groups based on meta-information: the SC group, ABSA group, and MAST group. Within each group, model merging is performed using an evolutionary algorithm. The MEM process integrates the output of each expert model within the group by weighted averaging. For each task $t$, the expert model $M_t$ is weighted by $w_t$, and the merged model $M_{merged,g}$ is constructed for the group. The weighted merging of expert models is expressed as follows.

$$M_{merged,g} = \sum_{t \in g} w_t M_t \quad (4)$$

where the weights $w_t$ are optimized using the evolutionary algorithm, with a focus on improving performance for the weak data $D_t$ of each task.

The evolutionary algorithm operates as follows: Initially, an initial population of multiple weight vectors is randomly generated, where each individual in the population represents a different combination of weights. Crossover is then performed, where the weight vectors of two parent individuals are combined to create new offspring. This allows for the exchange of genetic information between the parents and helps explore different regions of the search space. Following crossover, mutation occurs, where some weights are randomly altered to introduce additional variability and ensure the algorithm does not converge prematurely on suboptimal solutions. After crossover and mutation, models are merged, and inference is performed using the weak data $D_t$, after which fitness is evaluated. Fitness evaluation measures the performance of the merged model using metrics such as accuracy or F1 score. The fitness function $f(w_t)$ is defined as follows.

$$f(w_t) = \frac{1}{|D_t|}\sum_{i=1}^{|D_t|} \delta(\hat{O}_{t,i}, O_{t,i}) \quad (5)$$

where $\delta(\hat{O}_{t,i}, O_{t,i})$ represents the similarity between the model's prediction $\hat{O}_{t,i}$ and the ground truth label $O_{t,i}$, and the fitness score is calculated accordingly.

After the fitness evaluation, selection occurs, where individuals with higher fitness are chosen to propagate into the next generation. This process of crossover, mutation, and selection is repeated across multiple generations, ultimately optimizing the weights and producing the optimal merged model $M_{merged,g}$ for each group. The optimal merged models $M_{merged,SC}$, $M_{merged,ABSA}$, and $M_{merged,MAST}$ obtained for each group are further combined using an evolutionary model merging process to construct the final merged model $M_{final}$. This final weighted integration is expressed as follows.

$$M_{final} = \sum_g w_g M_{merged,g} \quad (6)$$

where $g$ represents each group (SC, ABSA, MAST), and the weights $w_g$ for each group are also optimized using the evolutionary algorithm. Through optimal weighting by the evolutionary algorithm, the final emotion-specialized LLM is completed, integrating all expert models. The algorithm for the merging phase is show in Fig. 5. In this way, MEM using weak data enables the model to overcome its weaknesses and achieve optimal weighting, leading to the completion of the sentiment-specialized LLM.

---

Algorithm: Model Merging Based on MEM
Input: Expert models $\{M_t\}$ for each sentiment analysis task $t$, weak data $D_t$ for each task, and the number of evolutionary generations
Output: Final merged model $M_{final}$

1: Initialize initial population of weight vectors $\{w_t\}$ randomly for each task $t$
2: **for** each group $g \in \{SC, ABSA, MAST\}$ **do**
3:   **for** each evolutionary generation **do**
4:     **for** each individual in the population **do**
5:       Merge expert models within the group using weighted sum
6:       $M_{merged,g} = \sum_{t \in g} w_t M_t$
7:       Perform inference on the weak data $D_t$ using $M_{merged,g}$
8:       Calculate fitness score $f(w_t)$ for each individual using:
9:       $f(w_t) = \frac{1}{|D_t|}\sum_{i=1}^{|D_t|} \delta(\hat{O}_{t,i}, O_{t,i})$
10:      where $\delta(\hat{O}_{t,i}, O_{t,i})$ measures the similarity between the predicted output $\hat{O}_{t,i}$ and the ground truth $O_{t,i}$
11:      Select individuals with higher fitness for the next generation
12:      Perform crossover and mutation to generate new weight vectors
13:     **end for**
14:   **end for**
15: **end for**
16: **for** each group $g \in \{SC, ABSA, MAST\}$ **do**
17:   Obtain the final merged model for the group $M_{final}$
18:   $M_{final} = \sum_g w_g M_{merged,g}$
19:   where $w_g$ is the optimized weight for each group as above
20: return final merged model $M_{final}$

Fig. 5. Algorithm for Model Merging Based on MEM.

### E. Curriculum Learning Based on ICL

MCL offers a systematic method for LLMs to adapt to a broad spectrum of sentiment analysis tasks by providing a learning sequence based on the difficulty of each task, as calculated from the task-specific meta-data. Specifically, we design a learning curriculum by quantitatively evaluating the difficulty of each task using meta-data extracted directly from the sentiment analysis tasks themselves. The difficulty of each task is determined by considering four key factors derived from the meta-data of the tasks: the number of sentiment classes ($C$), dataset diversity ($V$), task complexity ($Z$), and the subjective nature of the sentiment analysis task ($S$). As the number of sentiment classes increases, the complexity of the classification task also grows, reflecting the basic challenge in the classification process. A more diverse dataset requires the model to possess broader, more generalized knowledge, which increases the difficulty of handling the task. Task complexity is particularly significant in tasks such as aspect-based or subjective sentiment analysis, where deeper context understanding is required compared with simpler sentiment classification tasks. Subjective tasks, such as sarcasm detection or hate speech detection, pose additional challenges as they demand extensive contextual understanding and background knowledge, making them difficult even for humans. To quantify the difficulty, a score is assigned to each task and can be calculated as follows.

$$Score = w_1 \times C + w_2 \times V + w_3 \times Z + w_4 \times S \quad (7)$$

The weights $w_1$, $w_2$, $w_3$, $w_4$ are based on [22]. Specifically, the weight for the number of sentiment classes ($C$) is set to 1.0, reflecting the growing classification challenge as the number of classes increases. Dataset diversity ($V$) has a weight of 0.5, as it enhances generalization ability but does not impact difficulty as much as class count. Task complexity ($Z$) has a weight of 2.0, considering that complex tasks like aspect-based or subjective sentiment analysis demand higher context understanding. Lastly, subjective sentiment analysis ($S$) carries the highest weight of 5.0, as tasks like sarcasm or hate speech detection require significant cognitive effort.

Introducing MCL in the inference phase allows the LLM to progress from simpler tasks to more complex ones, efficiently utilizing its learned knowledge to improve performance across a wide range of sentiment analysis tasks. The curriculum learning process is shown in Fig. 6.

---

Algorithm: Curriculum Learning Based on ICL
Input: Sentiment analysis tasks $T = \{t_1, t_2, \ldots, t_k\}$, meta-information for each task, weights $w_1, w_2, w_3, w_4$
Output: Trained model with tasks processed in the order of difficulty

1: **for** each task $t \in T$ **do**
2:   Compute the difficulty score for the task $t$ using the following formula
3:   $Score = w_1 \times C + w_2 \times V + w_3 \times Z + w_4 \times S$
4: **end for**
5: **sort** the tasks $T$ in ascending order based on their difficulty scores
6: **combine** the instruction, input, and output for each sorted task into $X$
7: Create prompt $P_{t,i} = Prompt(R_t, I_{t,i}, A_{t,i}, X)$
8: Feed the prompt $P$ into the model
9: $\hat{O}_{t,i} = M_{final}(P_{t,i}; \theta_{final})$
10: return the output $\hat{O}$

Fig. 6. Algorithm for Curriculum Learning Based on ICL.

## IV. EXPERIMENT AND ANALYSIS

### A. Dataset and Experiment Design

In this study, multiple publicly available datasets, as shown in Table I, are used to conduct experiments, covering a wide range of sentiment analysis tasks. These datasets are sourced from various origins, underwent preprocessing, and are organized into final datasets. The data is split into training and test sets, with the SC tasks adjusted to maintain label balance. For testing, a maximum of 500 randomly selected samples per task are used for efficient evaluation. Accuracy, Macro-F1, and Micro-F1 scores are used as evaluation metrics.

TABLE I.    NUMBER OF DATA AND METRICS FOR EACH TASK

| Task | Train | Test | Metric |
|---|---|---|---|
| *Sentiment Classification (SC)* | | | |
| SC-2class-sen [24] | 6000 | 500 | Accuracy |
| SC-2class-doc [25, 26] | 3000 | 500 | Accuracy |
| SC-3class [27] | 3000 | 500 | Accuracy |
| SC-5class [26, 28] | 15000 | 500 | Accuracy |
| *Aspect-Based Sentiment Analysis (ABSA)* | | | |
| ABSA-ATSA [29] | 1600 | 421 | Accuracy |
| ABSA-ACSA [30] | 3000 | 400 | Accuracy |
| ABSA-TSD [29] | 1600 | 421 | Micro-F1 |
| ABSA-ASD [30] | 1500 | 400 | Micro-F1 |
| ABSA-ASTE [31] | 1500 | 500 | Micro-F1 |
| ABSA-ASQP [31] | 1500 | 500 | Micro-F1 |
| *Multifaceted Analysis of Subjective Text (MAST)* | | | |
| MAST-11class [32] | 5000 | 500 | Accuracy |
| MAST-28class [33] | 10000 | 500 | Accuracy |
| MAST-Hate [34] | 9000 | 500 | Macro-F1 |
| MAST-Offensive [34] | 3000 | 500 | Macro-F1 |
| MAST-Irony [35] | 3000 | 500 | Macro-F1 |

To evaluate the effectiveness of the proposed MEM-MCL model, we conduct experiments in two stages:
i) Evaluating the performance of MEM. Specifically, inference results from 15 sentiment analysis tasks are used to compare the model merged through MEM with the base model and individual expert models, which are created using Llama 2. The goal is to demonstrate the standalone effectiveness of MEM as a method for merging expert models through an evolutionary process, with a focus on whether it enhances performance across diverse tasks.
ii) Evaluating the performance of MEM integrated with MCL. In this stage, tasks are presented to the model based on their difficulty, which is defined using task-specific meta-data. The model is sequentially fed tasks, progressing from simpler to more complex ones. This approach aims to evaluate how MCL, which organizes the learning process based on the difficulty of sentiment analysis tasks, and enhances the model's task understanding and adaptability during inference.

In addition, Table II presents the key hyperparameter settings used throughout experiments. These are designed to measure the contributions of the MEM and MCL approaches in building a robust, generalized model that improves performance across a broad range of sentiment analysis tasks.

TABLE II.    HYPERPARMETER SETTINGS

| Phase | Parameter | Value |
|---|---|---|
| SFT phase | Batch size | 64 |
| | Micro batch size | 4 |
| | Learning rate | 3e-4 |
| | Epochs (Max) | 10 |
| | Weight decay | 0.1 |
| | LoRA r | 4 |
| | LoRA α | 16 |
| | LoRA Dropout rate | 0.05 |
| Merge phase | Evolutionary generations | 10 |
| | Initial population size | 20 |
| | Crossover point | Random |
| | Mutation probability | 0.1 |

### B. Performance on Multi-Stage Evolutionary Model Merging

The inference results for 15 sentiment analysis tasks are used to compare the model merged through MEM with the base model and individual expert models. The inference results are shown in Fig. 7.

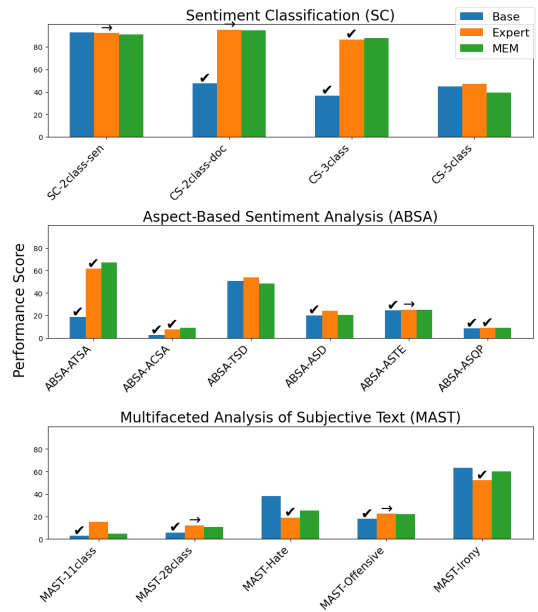

Fig. 7. Performance of MEM Compared with Other Models for Sentiment Analysis Tasks. A check mark (✔) indicates that the MEM model outperformed the comparison model in terms of performance score. Meanwhile, a rightward arrow (→) indicates that although the MEM model underperformed, the difference was within 1.6%, which corresponds to a difference of fewer than 10 incorrect predictions. The same notation applies to the following figures as well.

As shown in Fig. 7, the merge model obtained through MEM demonstrated outstanding performance better than the base model across many tasks. Notably, substantial improvements were observed in tasks like ATSA, ACSA, ASQP, and SC-3class. These tasks belong to the SC and ABSA groups, where the effects of MEM are particularly pronounced. Additionally, in the SC-2class-doc task, there is a notable performance improvement compared with the base model. This demonstrates that the MEM model consistently outperforms the base model, confirming the effectiveness of this multi-stage evolutionary merging approach as a single unified model. Next, when comparing the expert models with the MEM model, it is noteworthy that the MEM model performed at least as well as, and often better than, the expert models across most tasks. The expert models, which are fine-tuned specifically for each task, are optimized to deliver the best performance for their respective tasks. However, the fact that the MEM model, capable of supporting all tasks, demonstrated equal or superior performance, especially in tasks like ATSA, ACSA, and ASQP, further supports the powerful integration capability of the evolutionary model merging process. The effectiveness of the MEM model, which maintains high performance while handling multiple tasks, should be particularly emphasized.

Overall, the MEM model consistently outperformed the base model, confirming its robustness as a unified approach. Its superior performance in specific tasks compared with expert models further highlights the effectiveness of the evolutionary algorithm, demonstrating MEM's capability to manage multiple sentiment analysis tasks efficiently.

### C. Performance on Meta Data Driven Curriculum Learning

Before conducting the MCL experiments, the difficulty of each task is calculated based on four meta data-driven factors: $C, V, Z, S$. The results of this analysis are shown in Table III. Tasks are then ranked by difficulty according to these factors, and in the MCL phase, the learning curriculum is structured to progress from easier tasks to more difficult ones.

TABLE III. TASK DIFFICULTY AND RANKINGS BASED ON META-DATA

| Task | C | V | Z | S | Score | Rank |
|---|---|---|---|---|---|---|
| *Sentiment Classification (SC)* | | | | | | |
| SC-2class-sen | 2 | 1 | 1 | 0 | 4.5 | 1 |
| SC-2class-doc | 2 | 2 | 1 | 0 | 5 | 2 |
| SC-3class | 3 | 1 | 1 | 0 | 5.5 | 3 |
| SC-5class | 5 | 2 | 1 | 0 | 8 | 6 |
| *Aspect-Based Sentiment Analysis (ABSA)* | | | | | | |
| ABSA-ATSA | 2 | 1 | 1 | 0 | 4.5 | 4 |
| ABSA-ACSA | 3 | 1 | 1 | 0 | 5.5 | 5 |
| ABSA-TSD | 2 | 1 | 2 | 0 | 6.5 | 7 |
| ABSA-ASD | 3 | 1 | 2 | 0 | 7.5 | 8 |
| ABSA-ASTE | 3 | 1 | 3 | 0 | 9.5 | 12 |
| ABSA-ASQP | 3 | 1 | 4 | 0 | 11.5 | 13 |
| *Multifaceted Analysis of Subjective Text (MAST)* | | | | | | |
| MAST-11class | 11 | 1 | 1 | 1 | 18.5 | 14 |
| MAST-28class | 28 | 1 | 1 | 1 | 35.5 | 15 |
| MAST-Hate | 2 | 1 | 1 | 1 | 9.5 | 9 |
| MAST-Offensive | 2 | 1 | 1 | 1 | 9.5 | 10 |
| MAST-Irony | 2 | 1 | 1 | 1 | 9.5 | 11 |

To evaluate the effectiveness of MCL, three patterns of model performance are compared. The first is the result from the MEM model (MEM). The second involves the MEM model learning tasks in an unsystematic order (Unstructured), and the third uses MEM with MCL, where tasks are learned in order of increasing difficulty (MEM-MCL). The inference results for each task across these three patterns are shown in Fig. 8.

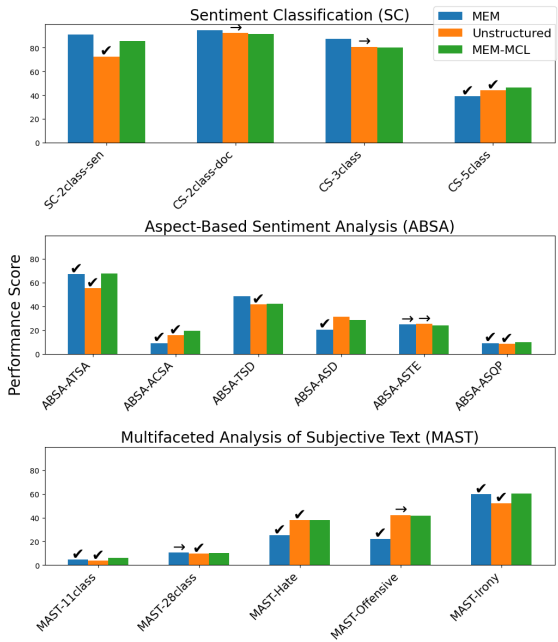

Fig. 8. Performance of MEM-MCL Compared with Other Methods for Sentiment Analysis Tasks.

When comparing the results of applying MCL with those of the MEM model, MEM-MCL outperforms MEM in more than half of the tasks. Particularly, significant improvements are observed in ACSA and ASD, confirming the effectiveness of a curriculum based on task difficulty. MEM-MCL also surpasses the MEM in the MAST-Hate task, further demonstrating its advantages. Additionally, when compared with the results of learning tasks in an Unstructured, MEM-MCL performs equally well or better in many tasks. Notably, tasks such as ATSA and MAST-Irony showed significant improvements, indicating that learning tasks in a structured order of difficulty enhances model performance. However, in some tasks, particularly those in the SC group, MEM outperforms the MEM-MCL, and in a few instances, such as in the ASD task, unstructured learning order outperforms MEM-MCL. These results indicate that the impact of MCL may be limited in certain tasks, highlighting the need for further investigation and refinement of the curriculum. In summary, MCL is proved effective across a variety of tasks, particularly by applying a curriculum based on difficulty rankings derived from task meta data, which allows for more seamless knowledge extraction and thus can improve the overall performance of the model.

*D. Discussion*

As mentioned above, we evaluate the effectiveness of our proposed MEM-MCL, which leverages meta data from sentiment analysis tasks and addresses weak data. The results show that the merged model using MEM achieves accuracy comparable to or exceeding that of expert models created through instruction tuning across multiple sentiment analysis tasks, even when used as a standalone model. Additionally, MCL facilitates knowledge extraction and improves overall accuracy. These findings indicate that the proposed model offers a promising new way for sentiment analysis in large language modeling.

However, there are still several challenges. First, certain tasks presented difficulties with instruction tuning. For instance, in tasks with significant label imbalances, e.g., MAST-Hate, the small 7B model sometimes struggled to control output formats. This may because of both the model's size and the complexity of the task. Additionally, model merging effects are not uniformly prominent, with some tasks showing lower accuracy than the base model. For more complex or subjective tasks, the weighting or algorithm adjustments during merging may be insufficient. Future research could focus on refining these parameters, potentially allowing accuracy to surpass the base model across all tasks. Regarding MCL, ranking tasks by difficulty using meta data was effective, but optimizing the weight assignments remains challenging. In our model, weights are set heuristically based on prior research but may not be fully optimized. Future research should explore methods for automatically determining optimal weights in MCL. A dynamic mechanism that adjusts weights based on the model's performance in individual tasks could be a promising direction for improvement.

V. CONCLUSION

In this paper, we developed a hybrid model capable of handling a wide range of sentiment analysis tasks by the proposed MEM-MCL model. Traditional sentiment analysis models are typically specialized for individual tasks, making it challenging to manage multiple tasks simultaneously. The

MEM enhanced the accuracy of an integrated model using evolutionary algorithms, while MCL was introduced to guide learning sequences based on task difficulty. Experiment results demonstrated that the MEM model significantly outperformed the base model in various tasks, particularly in ABSA and MAST tasks. While the MCL facilitated more effective learning process, resulting in further accuracy improvements.

In future studies, building upon the current achievements, more focus will be placed on optimizing the evolutionary algorithm and automating the MCL process to enhance accuracy. These improvements will further refine the model's performance, particularly in more complex sentiment analysis tasks, and enable practical applications in real-world scenarios.

ACKNOWLEDGMENT

This work was supported in part by the Grants-in-Aid for Scientific Research (C) from Japan Society for the Promotion of Science (JSPS) under Grant 23K11064, and in part by the Japan Safety Society Research Center (JSSRC).

REFERENCES

[1] J. R. Jim, M. A. R. Talukder, P. Malakar, M. M. Kabir, K. Nur, and M. F. Mridha, "Recent advancements and challenges of NLP-based sentiment analysis: A state-of-the-art review," *Natural Language Processing Journal*, vol. 6, p. 100059, Mar. 2024. doi: 10.1016/j.nlp.2024.100059.

[2] Q. Zhong, L. Ding, J. Liu, B. Du, and D. Tao, "Can ChatGPT Understand Too? A Comparative Study on ChatGPT and Fine-tuned BERT," *arXiv*, Feb. 2023.

[3] Z. Wang, Q. Xie, Y. Feng, Z. Ding, Z. Yang, and R. Xia, "Is ChatGPT a Good Sentiment Analyzer? A Preliminary Study," *arXiv*, Apr. 2023.

[4] B. Zhang, H. Yang, and X.-Y. Liu, "Instruct-FinGPT: Financial sentiment analysis by instruction tuning of general-purpose large language models," *SSRN Electron. J.*, Jun. 2023. doi: 10.2139/ssrn.4489831.

[5] T. Akiba, M. Shing, Y. Tang, Q. Sun, and D. Ha, "Evolutionary optimization of model merging recipes," *arXiv*, Mar. 2024.

[6] A. Vaswani et al., "Attention Is All You Need," *arXiv*, Jun. 2017.

[7] M. Reid et al., "Gemini 1.5: Unlocking multimodal understanding across millions of tokens of context," *arXiv*, Mar. 2024.

[8] H. Touvron et al., "Llama 2: Open Foundation and Fine-Tuned Chat Models," *arXiv*, Jul. 2023.

[9] S. Bubeck et al., "Sparks of Artificial General Intelligence: Early experiments with GPT-4," *arXiv*, Mar. 2023.

[10] X. Deng, V. Bashlovkina, F. Han, S. Baumgartner, and M. Bendersky, "LLMs to the Moon? Reddit Market Sentiment Analysis with Large Language Models," in *Companion Proceedings of the ACM Web Conference 2023*, pp. 1014–1019. Apr. 2023. doi: 10.1145/3543873.3587605.

[11] X. Jin, X. Ren, D. Preotiuc-Pietro, and P. Cheng, "Dataless knowledge fusion by merging weights of language models," *arXiv*, Dec. 2022.

[12] M. Li et al., "Branch-train-Merge: Embarrassingly parallel training of expert language models," *arXiv*, Aug. 2022.

[13] S. Xiao, Z. Liu, P. Zhang, and X. Xing, "LM-Cocktail: Resilient Tuning of Language Models via Model Merging," in *Findings of the Association for Computational Linguistics ACL 2024*, pp. 2474–2488. Aug. 2024.

[14] M. Taboada, J. Brooke, M. Tofiloski, K. Voll, and M. Stede, "Lexicon-Based Methods for Sentiment Analysis," *Comput. Linguist.* vol. 37, pp. 267-307, Jun. 2011. doi: 10.1162/COLI_a_00049.

[15] T. Abdullah and A. Ahmet, "Deep learning in sentiment analysis: Recent architectures," *ACM Comput. Surv.*, vol. 55, no. 8, pp. 1–37, Aug. 2023. doi: 10.1145/3548772.

[16] Z. Wang, V. J. C. Tong, and H. C. Chin, "Enhancing Machine-Learning Methods for Sentiment Classification of Web Data," in *Information Retrieval Technology*, pp. 394–405. 2014. doi: 10.1007/978-3-319-12844-3_34.

[17] A. Mohammadi and A. Shaverizade, "Ensemble deep learning for aspect-based sentiment analysis," *International Journal of Nonlinear Analysis and Applications*, vol. 12, no. Special Issue, pp. 29–38, Dec. 2021. doi: 10.22075/IJNAA.2021.4769.

[18] K. He, R. Mao, T. Gong, C. Li, and E. Cambria, "Meta-based self-training and re-weighting for aspect-based sentiment analysis," *IEEE Trans. Affect. Comput.*, vol. 14, no. 3, pp. 1731–1742, Jul. 2023. doi: 10.1109/taffc.2022.3202831.

[19] J. Sido and M. Konopík, "Curriculum Learning in Sentiment Analysis," in *Speech and Computer*, 2019, pp. 444–450. doi: 10.1007/978-3-030-26061-3_45.

[20] V. A. Rao, K. Anuranjana, and R. Mamidi, "A Sentiwordnet Strategy for Curriculum Learning in Sentiment Analysis," *Springer International Publishing*, pp. 170–178, 2020. doi: 10.1007/978-3-030-51310-8_16.

[21] A. Dahiya, N. Battan, M. Shrivastava, and D. M. Sharma, "Curriculum Learning Strategies for Hindi-English Code-Mixed Sentiment Analysis," in *Artificial Intelligence. IJCAI 2019 International Workshops*, pp. 177–189, Aug. 2020. doi: 10.1007/978-3-030-56150-5_9.

[22] W. Zhang, Y. Deng, B. Liu, S. Pan, and L. Bing, "Sentiment Analysis in the Era of Large Language Models: A Reality Check," *Association for Computational Linguistics*, pp. 3881–3906, Jun. 2024. doi: 10.18653/v1/2024.findings-naacl.246.

[23] E. J. Hu et al., "LoRA: Low-Rank Adaptation of Large Language Models," *arXiv*, Jun. 2021.

[24] R. Socher et al., "Recursive Deep Models for Semantic Compositionality Over a Sentiment Treebank," in *Proceedings of the 2013 Conference on Empirical Methods in Natural Language Processing*, 2013, pp. 1631–1642. Oct. 2013.

[25] A. L. Maas, R. E. Daly, P. T. Pham, D. Huang, A. Y. Ng, and C. Potts, "Learning Word Vectors for Sentiment Analysis," in *Proceedings of the 49th Annual Meeting of the Association for Computational Linguistics: Human Language Technologies*, pp. 142–150. Jun. 2011.

[26] X. Zhang, J. Zhao, and Y. LeCun, "Character-level Convolutional Networks for Text Classification," *arXiv*, Sep. 2015.

[27] Y. Hou, J. Li, Z. He, A. Yan, X. Chen, and J. McAuley, "Bridging language and items for retrieval and recommendation," *arXiv [cs.LG]*, vol. abs/2403.03952, Mar. 2024.

[28] "SetFit/sst5 · Datasets at Hugging Face." [Online]. Available: https://huggingface.co/datasets/SetFit/sst5. [Accessed: 07-Aug-2024].

[29] uclnlp/jack/data/sentihood. Github, [Accessed: 07-Aug-2024].

[30] Aspect_Based_Sentiment_Analysis. Github, [Accessed: 07-Aug-2024].

[31] W. Zhang, Y. Deng, X. Li, Y. Yuan, L. Bing, and W. Lam, "Aspect Sentiment Quad Prediction as Paraphrase Generation," in *Proceedings of the 2021 Conference on Empirical Methods in Natural Language Processing*, pp. 9209–9219. Nov. 2011. doi: 10.18653/v1/2021.emnlp-main.726.

[32] S. Mohammad, F. Bravo-Marquez, M. Salameh, and S. Kiritchenko, "SemEval-2018 Task 1: Affect in Tweets," in *Proceedings of The 12th International Workshop on Semantic Evaluation*, New Orleans, Louisiana, Jun. 2018. doi: 10.18653/v1/S18-1001.

[33] D. Demszky, D. Movshovitz-Attias, J. Ko, A. Cowen, G. Nemade, and S. Ravi, "GoEmotions: A Dataset of Fine-Grained Emotions," in *Proceedings of the 58th Annual Meeting of the Association for Computational Linguistics*, pp. 4040–4054. Jul. 2020. doi: 10.18653/v1/2020.acl-main.372.

[34] "hs-knowledge/hateval_enriched · Datasets at Hugging Face." [Online]. Available: https://huggingface.co/datasets/hs-knowledge/hateval_enriched. [Accessed: 07-Aug-2024].

[35] O. Rohanian, S. Taslimipoor, R. Evans, and R. Mitkov, "WLV at SemEval-2018 Task 3: Dissecting Tweets in Search of Irony," in *Proceedings of the 12th International Workshop on Semantic Evaluation*, pp. 553–559. Jun. 2018. doi: 10.18653/v1/S18-1090.